# Shared Context Candidate Scoring with Jointly Adapted Readout Networks

Jingyang Du[1,†], Yiben Fu[†], Yang Shen[2,3,*]

[1]School of Journalism and Communication, Minzu University of China, Beijing, China
[2]School of Journalism and Communication, Tsinghua University, Beijing, China
[3]College of AI, Tsinghua University, Beijing, China

[†]These authors contributed equally to this work and share first authorship.
[*]Corresponding author: Yang Shen (124739259@qq.com)
ORCID: Yang Shen 0000-0003-4814-9018

**Abstract**

Repeated decisions over a shared document create two distinct opportunities: reusing document computation and adapting how question representations score candidate answers. We present Q64, a candidate-scoring system that preserves a frozen 64-layer Qwen3.8-27B text backbone, reuses its hybrid prefix state and intermediate document representations, and jointly trains a residual adapter with a four-layer decoder and candidate-conditioned readout. We used ZiYor, a self-evolving AI system, to explore and refine the Q64 design. We specified the scoring method, implemented the system, and evaluated its accuracy and latency. The final adaptation stage updates 88.66 million parameters using 8,192 selected training views. On a fixed 5,112-item evaluation combining ARC-Challenge, ANLI-R3, and stratified subsets of MMLU-Pro and CMMLU, joint adaptation improves four-benchmark macro accuracy from 66.46% to 74.32%. It remains below the native 27B baseline at 75.68% and AnyJev Qwen3-32B L0 at 76.68%. On a separate synthetic workload with 4,096 document tokens and 50 questions, median cold-request latency is 2.237 seconds, compared with 5.383 seconds for the native baseline and 56.872 seconds for the measured batched AnyJev configuration. Short single-question requests do not share this advantage. The evidence supports an accuracy-latency tradeoff in the evaluated systems, rather than a backbone reduction or an isolated component advantage. Public benchmarks informed earlier development, and real-document multi-question validation remains open.

## 1 Introduction

A single document can support many decisions: assigning labels, checking claims, or selecting among task-specific alternatives. Repeatedly invoking a language model for each decision can duplicate document computation. A reusable document state addresses this duplication, but does not by itself specify how the model should compare changing candidate texts. A second design choice is therefore the readout interface between language-model representations and the final decision distribution.

This paper studies these choices together in Q64, an implemented system for shared-context candidate scoring. A frozen language-model backbone supplies both document representations and full-depth question representations. A trainable adapter, four-layer decoder, and candidate-conditioned reader transform these representations into scores over the supplied candidates. The label set can change between requests because the system scores candidate text rather than using a fixed task-specific classifier. Q64 denotes a 64-layer question path; it does not denote quantization or a 64-parameter model.

The study addresses two empirical questions. First, can joint adaptation of an existing readout improve decision accuracy without extending its inference architecture? Second, when do document reuse and independent question batching reduce the latency of the complete scoring request? These questions require separate evidence. The accuracy evaluation uses public benchmark items, whereas latency is measured on controlled synthetic shapes. Neither experiment establishes that the accuracy and latency figures are achieved simultaneously on a production workload.

The contribution is an integration and empirical characterization of shared hybrid prefix states, deep question representations, and jointly adapted candidate scoring. We describe the implementation at the level of states, dimensions, training scope, and cache validity; reconstruct nine systems' accuracy from archived per-item predictions; and analyze six latency configurations from raw timing repetitions. The resulting picture is selective: joint adaptation improves the parent model substantially, but the system retains an accuracy deficit to the two strongest comparisons and incurs cold-start overhead for short single-question requests. Existing work already supplies prefix sharing, evidence fusion, and bottleneck adaptation, so we do not claim these components individually as new.

## 2 Related work

Prefix reuse and serving. PagedAttention manages attention key-value storage in blocks and supports memory sharing between sequences [1]. Its shared-prefix use case already permits precomputed prefix states to be reused across requests. Q64 addresses a particular hybrid backbone whose prefix comprises attention keys and values as well as convolutional and recurrent states. It additionally retains document feature banks for a trained reader. These implementation choices do not establish a general advantage over PagedAttention or optimized serving engines, which were not evaluated here.

Evidence fusion. Fusion-in-Decoder independently encodes retrieved passages together with the question, then lets a decoder attend across the combined representations [2, Section 3]. Q64 instead reuses a question-independent document encoding across several questions and produces a distribution over candidate texts. This changes the input-sharing contract and output interface. Cross-attention over document evidence itself is established prior work, and this study does not compare Q64 against Fusion-in-Decoder on a matched task.

Parameter-efficient adaptation. Residual bottleneck adapters provide a small trainable transformation around pretrained representations [3]. Q64 uses this established design at the input of a separately trained readout network. The final adaptation stage updates the existing decoder, document projections, pointer, and candidate reader together with the adapter. It therefore differs from an adapter-only update, but the recorded comparison also changes training coverage; it cannot isolate the causal contribution of unfreezing any one module.

Direct decision interfaces. AnyJev exposes candidate decisions through an existing language model [4]. The evaluated L0 configuration reads label logits and averages over the complete set of cyclic candidate rotations. Its Hugging Face backend can share token-exact prefixes, including within the scored rotation group [5]. The recorded Q64 comparison changes the backbone, readout, rotation policy, precision, and batching strategy together. It is a system comparison, not proof that direct readout is intrinsically faster than every possible AnyJev deployment.

# 3 Shared context candidate scoring

## 3.1 Task and state contract

Let a request contain one document D and Q questions. Question j has a text $x_j$ and $K_j$ candidate texts. The output is one normalized score vector per question and its highest-scoring candidate. The system does not generate an explanatory answer. Probabilities are normalized model scores; calibration and abstention quality are not established by the present experiments.

$$p_{jk} = \frac{\exp(s_{jk})}{\sum_{r=1}^{K_j} \exp(s_{jr})}, \qquad \hat{y}_j = \arg\max_k p_{jk} \tag{1}$$

All questions in a request must share the same document. Backbone weights, tokenization, candidate encoding, and readout parameters must remain compatible with the cached state. The measured reader checks document-token identity before warm reuse and validates the frozen-prefix producer when acquiring a prefix. Cache validity assumes fixed model weights and tokenization; changing the backbone or readout requires explicit invalidation and rebuilding. Questions receive separate branch views of the shared prefix and do not attend to each other. Within a question, however, candidates are encoded in a causal sequence. Candidate independence and permutation invariance therefore do not follow from cross-question isolation.

## 3.2 Document encoding and hybrid prefix reuse

The backbone is the text component of Qwen3.8-27B, with 64 layers and hidden width 5,120 [6]. It interleaves Gated DeltaNet and standard gated attention. The entire text backbone is retained and frozen; the vision encoder is unused. Document encoding produces two kinds of reusable objects: the hybrid prefix state needed to continue the backbone, and token-level feature banks from layers 32 and 64 needed by the readout.

The prefix contains attention keys and values together with the convolutional and recurrent states of the linear-attention layers. The implementation caches the prefix at a 64-token boundary. A residual document tail is replayed before each question branch, so reuse does not imply that every document token is cached at every boundary. The branch view shares attention tensors and copies recurrent states when required by the batch or kernel mutation behavior. Same-length question branches are grouped in batches of at most 16.

The document feature banks are denoted M32 and M64. The final-layer bank includes the backbone's final normalization, while the intermediate bank is captured at layer 32. Question and candidate tokens continue through all 64 frozen backbone layers to obtain their final normalized representations. The document state is therefore amortized across questions, but each question still incurs a full-depth continuation. This is not a shallow replacement for the backbone.

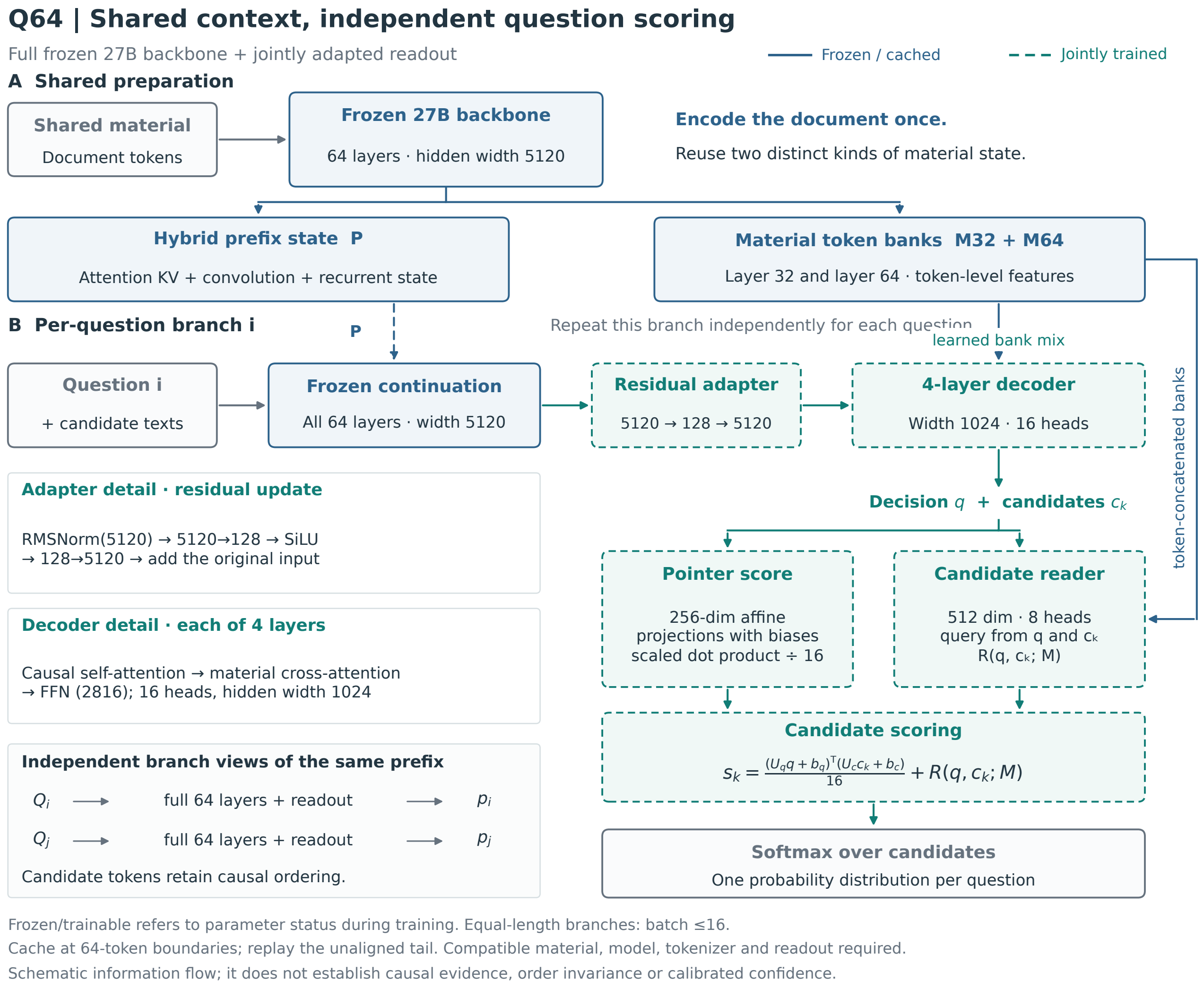

Figure 1. Q64 inference schematic. (A) Frozen document computation supplies a hybrid continuation state and two token-level feature banks. (B) Independent question branches pass through the full frozen backbone, then a jointly trained residual adapter and four-layer decoder. The final score combines an affine pointer and candidate-conditioned document readout. Dimensions, cache boundaries, and branch batching are shown explicitly. The drawing specifies information flow, not measured causal attribution.

### 3.3 Residual adaptation and the four-layer decoder

For a question-path hidden representation H, the adapter first normalizes the 5,120-dimensional input, projects it to a bottleneck of width 128, applies SiLU, and projects back with a residual connection:

$$\widetilde{H} = H + W_u \,\mathrm{SiLU}(W_d \,\mathrm{RMSNorm}(H)) \tag{2}$$

The adapted representations are normalized and projected to width 1,024. Four decoder blocks then apply causal self-attention, cross-attention to a shared document memory, and a gated feed-forward network. Each block has 16 attention heads and feed-forward width 2,816. The source names this module S4 solely because it has four layers; we use four-layer decoder to avoid confusion with state-space sequence models.

For decoder cross-attention, each document bank has its own normalization and a learned scalar mixture weight. The mixture is projected once to decoder width, and each decoder layer prepares its own memory keys and values. In schematic notation, with the two mixture weights normalized by softmax:

$$\overline{M} = \sum_{\ell \in \{32, 64\}} \alpha_\ell \,\mathrm{RMSNorm}_\ell(M_\ell), \qquad \alpha = \mathrm{softmax}(a) \tag{3}$$

At inference, the projected memories are reusable for a fixed document and fixed readout weights. During adaptation, only raw frozen backbone features are cached. Trainable document mixtures and projections are recomputed with gradients on every training forward; otherwise the cached tensors would be stale and would sever the intended update path.

### 3.4 Candidate-conditioned readout

The decoder provides a decision-position vector q and a terminal vector $c_k$ for each candidate. A pointer uses two affine maps to dimension 256 and computes their scaled dot product, including the learned biases. A separate reader normalizes q and $c_k$ and projects them to dimension 512. Their sum supplies an eight-head attention query over the concatenated layer-32 and layer-64 document banks. Here MHA means concatenated scaled-dot-product attention heads without a learned output projection, followed by the displayed normalization. Each bank has its own normalization and shares a key-value projection. Unlike the decoder mixture, this reader retains both banks as separate attention positions.

$$\hat{q} = P_q \, \mathrm{RMSNorm}(q), \quad \hat{c}_k = P_c \, \mathrm{RMSNorm}(c_k) \tag{4}$$

$$m_k = \mathrm{RMSNorm}(\mathrm{MHA}(W_a(\hat{q} + \hat{c}_k), K_M, V_M)) \tag{5}$$

A shared two-layer MLP with GELU scores the concatenation of the retrieved vector, projected decision and candidate vectors, and elementwise interactions. The final score is:

$$s_k = \frac{(U_q q + b_q)^\top (U_c c_k + b_c)}{16} + g([m_k, \hat{q}, \hat{c}_k, m_k \odot \hat{c}_k, \hat{q} \odot \hat{c}_k, |m_k - \hat{c}_k|]) \tag{6}$$

Invalid padded candidates receive negative-infinite logits before normalization. This readout exposes a route by which candidate text can select document information, but the existence of attention weights does not demonstrate that predictions depend on the correct evidence. Material interventions and candidate-order sensitivity require separate evaluation.

### 3.5 Cost boundary

A schematic cost decomposition separates document preparation from the grouped question computation. For a cold request, latency comprises frozen document encoding, readout-memory preparation, and all question batches. A warm request starts with compatible document states and prepared readout memory. If E denotes document encoding, P denotes readout preparation, $R_b$ denotes question-batch computation, and O denotes request bookkeeping and result assembly, the measured boundary can be represented as:

$$T_{\mathrm{cold}} = T_E + T_P + \sum_b T_{R_b} + T_O, \quad T_{\mathrm{warm}} = \sum_b T_{R_b} + T_O' \tag{7}$$

Each $R_b$ includes full-depth backbone continuation, residual-tail replay where needed, the readout network, and returning probabilities to the CPU. The overhead terms include grouping, validity checks, and result assembly and need not match across cold and warm requests. The terms were not separately timed; this is a schematic description rather than an asymptotic speedup theorem. Runtime depends on token lengths, candidate counts, batching, kernel implementation, and memory capacity. The measurements exclude model loading and CPU tokenization or prompt preparation.

## 4 Training protocol

### 4.1 Initialization and update scope

The final stage starts from the historical FI checkpoint's semantic modules and a development-selected Q64 input adapter at step 64. FI denotes the earlier training variant in the supplied experiment lineage. The backbone is frozen throughout the final stage, with no query-path LoRA and no knowledge-distillation objective. The adapter, decoder, document normalizers and projections, candidate reader, final pointer, and scorer are jointly updated using answer cross-entropy.

The selected incremental checkpoint stores 88,694,786 float32 parameters across 76 tensors. Of these, 32,768 belong to an unused frozen alignment component. The source-defined trainable count for the final stage is therefore 88,662,018, including 1,315,840 adapter parameters. These counts come from safetensors headers and the source update mask. They do not imply that inference requires only these parameters: the full frozen 27B backbone is also needed.

### 4.2 Training views and selection

Table 1. Composition of the final-stage training views. These are selected records from historical pools, not newly collected independent examples.

| Category | Views | Recorded composition |
| --- | --- | --- |
| Reading | 3,072 | BoolQ, C3, LogiQA |
| Inference and sentence pairs | 2,048 | OCNLI, AFQMC, MNLI variants |
| Classification | 1,536 | 12 historical source labels, 128 each |
| Natural-language rules | 512 | ProofWriter open-world subset |
| Paired rule families | 1,024 | 256 families, four conditions each |
| Total | 8,192 | One shuffled pass |

Each rule family contains a base example, a factual flip, an irrelevant change, and a paraphrase. These are related training views, not four independent observations. Each row receives one fixed random candidate permutation, with its target remapped accordingly. There is no auxiliary intervention or process loss in this final stage. Material hashes after Unicode normalization, group identifiers, and family identifiers exclude exact overlaps with the historical development and public evaluation pools. Such checks do not exclude semantic near-duplicates or overlap in the backbone's pretraining data.

The recorded run uses seed 2026100402, effective batch size 16, and 512 optimizer steps. AdamW uses learning rate 5e-5 for the adapter, 1e-5 for other trainable parameters, weight decay 0.01, 32 warmup steps, cosine decay, and gradient clipping at 1.0. Frozen feature tensors are stored in FP16 and updates use mixed precision. The protocol records 27.71 GiB of raw features and no training/development truncation, but the full historical training pools and feature cache are not included in the supplied artifacts.

The development set contains 1,536 historical records: 704 natural-language items, 320 rule items, and 512 regression items. At steps 0, 128, 256, 384, and 512, the selection criterion combines source-macro accuracy within each cohort:

$$S_{\text{dev}} = 0.50A_{\text{natural}} + 0.25A_{\text{rules}} + 0.25A_{\text{regression}} \quad (8)$$

The recorded tie-breakers prefer lower negative log-likelihood and then earlier checkpoints; the initial checkpoint is eligible. Step 512 has the highest composite score, 74.93%, with NLL 0.5961, compared with 65.54% and 0.7639 at step 0. This development score is distinct from the public four-benchmark macro. Source code and protocol records specify the rule, but the present artifact audit is not an external preregistration of that rule.

The reported final-stage pipeline took about 58.76 minutes, including frozen-feature extraction, updates, public evaluation, and timing; feature extraction accounted for about 34.57 minutes. This omits earlier FI training, adapter development, and base-model training. It must not be interpreted as total training cost from the original backbone. The intervention changes both trainable scope and training coverage, so the resulting gain describes the whole adaptation procedure.

## 5 Evaluation protocol

### 5.1 Accuracy tasks and aggregation

Table 2. Public evaluation composition. Asterisks indicate fixed stratified subsets rather than full benchmark leaderboards.

| Benchmark | Items | Selection and task |
| --- | --- | --- |
| ARC-Challenge [7] | 1,172 | Complete test split; science multiple choice |
| ANLI-R3 [8] | 1,200 | Complete test_r3; adversarial inference |
| MMLU-Pro* [9] | 1,400 | 100 items per subject across 14 subjects |
| CMMLU* [10] | 1,340 | 20 items per subject across 67 subjects |
| Total | 5,112 | Identical recorded item IDs across nine systems |

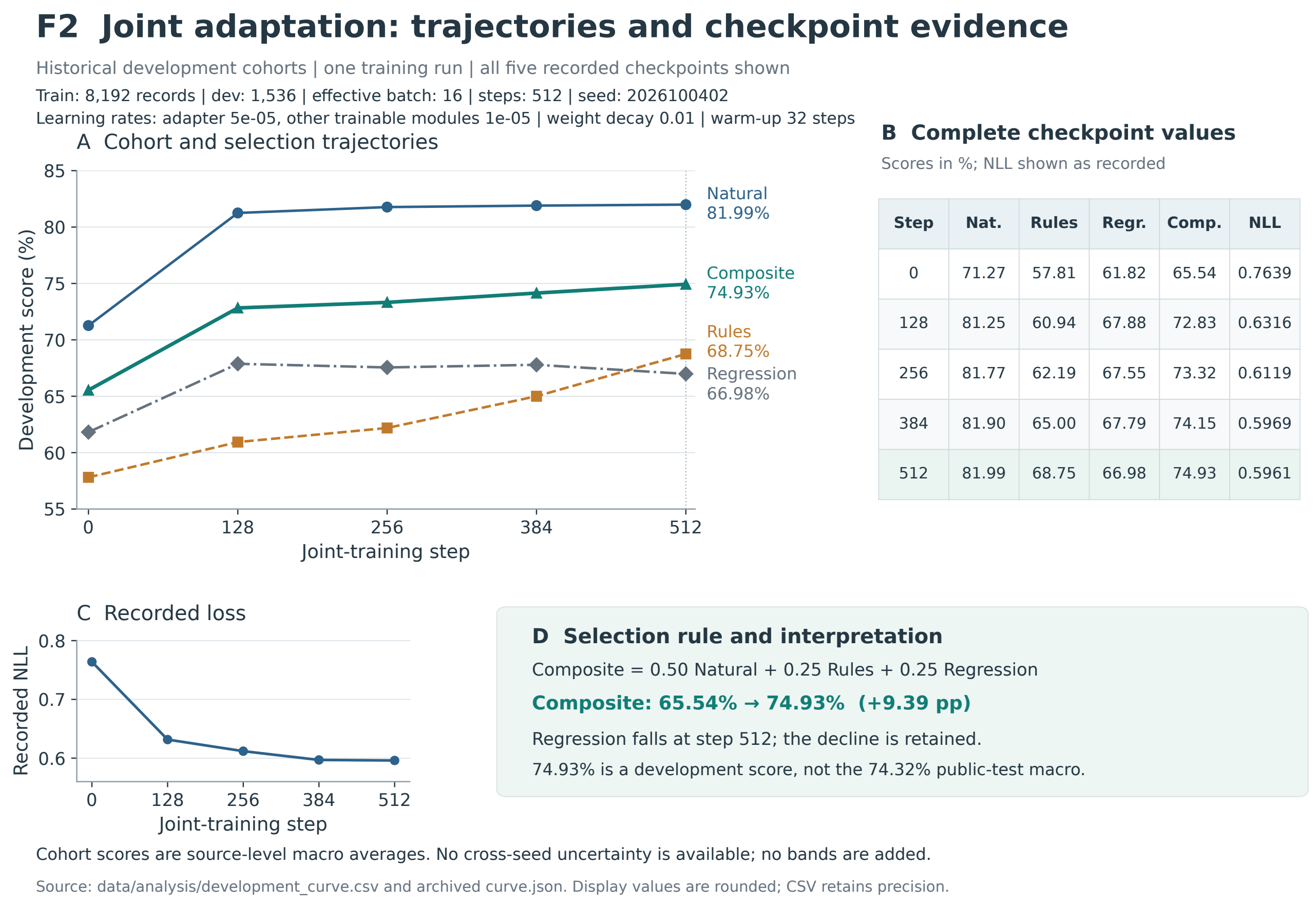

Figure 2. Recorded development results during the final joint-adaptation stage: (A) cohort and selection-score trajectories, (B) all five checkpoint values, (C) recorded negative log-likelihood (NLL), and (D) the selection rule and observed changes. Each cohort curve is a macro average over its historical sources; selection weights are 0.50, 0.25, and 0.25. These points evaluate one training run and are not public-test learning curves or estimates across random seeds.

Within each MMLU-Pro or CMMLU subject, the selection code sorts items by SHA256(seed/id) and takes the required count, with seed public-benchmark-v1/20261003. The model input excludes official answer and rationale fields. ARC, MMLU-Pro, and CMMLU put the benchmark question in the document field and use a short answer-selection instruction as the question branch; ANLI uses the premise as the document and the hypothesis in its branch. The accuracy study thus evaluates a common decision interface, not a corpus of 50-question real-document sessions.

For benchmark b, accuracy is first computed within each subject and then averaged equally over subjects. ARC and ANLI each use one subject group. The primary metric averages the four benchmark scores equally:

$$A_{\text{macro}} = \frac{1}{4} \sum_{b=1}^{4} \frac{1}{|S_b|} \sum_{s \in S_b} \frac{1}{n_{bs}} \sum_{i \in (b,s)} \mathbf{1}[\hat{y}_i = y_i] \tag{9}$$

Pooled item accuracy is also reported but answers a different question because benchmark sizes differ. The MMLU-Pro and CMMLU entries should not be compared directly with full-test, few-shot, or chain-of-thought leaderboard scores. The public item collection had already informed repeated diagnostics and research choices. Although the final checkpoint is selected using the historical development set, this public collection is not a fresh blind test.

### 5.2 Comparison systems

The main comparisons are native Qwen3.8-27B, AnyJev with Qwen3-32B L0, and the Q64 adapter-only parent. The native model disables explicit thinking and scores candidate labels in the original order. AnyJev uses the official implementation at commit e6efe1233a57ddc8b291eeb206142bcec003d10b, the Hugging Face backend, BF16, and complete cyclic candidate rotations. The 27B systems use FP16. Different backbones, templates, training histories, and precision prevent attributing all differences to the readout architecture.

Additional historical comparisons include a full-depth trained branch with LoRA and a pointer, a Q32 adapter-only branch, two Q16 variants (old A and FI), and the original Laya multilingual system. Their training histories and input constraints differ. The Laya record preserves full inputs on 5,065 of 5,112 items and truncates 47; its result represents that configuration, not its unconstrained performance. The Q16-to-Q64 sequence is a record of design development, not a controlled depth ablation.

### 5.3 Timing and numerical reanalysis

Latency records use one RTX PRO 6000 GPU, documents of 256 or 4,096 tokens, and requests with 1, 10, or 50 questions, each with eight candidates. Question identifiers differ but the workload is synthetic and controlled by shape. We report median complete-request time over nine repetitions. A cold request includes document encoding and state construction; a warm Q64 request reuses existing document state. Models and kernels are loaded and warmed before measurement.

The native reference already reuses material state but processes questions serially. Q64 groups same-length questions in batches up to 16. AnyJev batching flattens rotations from different questions into the official scoring backend: short inputs use rotation batches of 64, or eight questions; long inputs use 16, or two questions, after a batch of 32 exceeds available memory. These are capacity-selected settings, not an exhaustive search for each system's fastest deployment. Earlier native and serial AnyJev measurements and later batched AnyJev measurements are on the same card but were not interleaved.

For this manuscript, all 46,008 stored prediction rows are reanalyzed without rerunning model inference. The audit checks unique item IDs, targets, benchmark and subject identifiers, candidate counts, and stored predictions against probability argmax. Candidate strings and complete input texts are absent from the prediction logs; their ordering and exclusion of answer text are supported by code and protocol, not independently reconstructed forward inputs. All 624 available timing summaries are recomputed from their archived repetition arrays.

Post-hoc uncertainty analysis resamples paired correctness outcomes within each of the 83 benchmark-subject strata, preserving stratum sizes and the macro weighting. We use 20,000 bootstrap replicates and seeds 20261029 through 20261032 in the comparator order of Table 4, and report percentile 95% intervals for the macro difference. These intervals condition on the fitted models, chosen tasks, and item-resampling assumptions. They neither represent training-seed variability nor correct for benchmark feedback, and they do not cover all possible dependence between related items.

## 6 Results

### 6.1 Accuracy and paired changes

Table 3. Accuracy in percent on the same recorded items. MMLU-Pro* and CMMLU* are fixed subsets. Macro equally weights benchmarks after subject averaging; pooled accuracy equally weights items.

| System | ARC | ANLI | MMLU* | CMMLU* | Macro | Pooled |
|---|---|---|---|---|---|---|
| Q64 joint readout | 94.62 | 68.42 | 58.86 | 75.37 | 74.32 | 73.63 |
| Native 27B | 96.33 | 65.00 | 62.00 | 79.40 | 75.68 | 75.14 |
| AnyJev Qwen3-32B L0 | 95.65 | 65.08 | 62.64 | 83.36 | 76.68 | 76.21 |
| Full-depth trained branch | 93.77 | 57.25 | 58.07 | 72.24 | 70.33 | 69.78 |
| Q64 adapter only | 92.41 | 45.00 | 55.21 | 73.21 | 66.46 | 66.06 |
| Q32 adapter only | 86.60 | 49.58 | 43.71 | 63.28 | 60.80 | 60.05 |
| Old A Q16 | 54.01 | 38.83 | 25.57 | 42.69 | 40.28 | 39.69 |
| FI Q16 | 54.10 | 33.50 | 26.14 | 45.15 | 39.72 | 39.26 |
| Laya multilingual | 28.24 | 35.92 | 11.43 | 29.25 | 26.21 | 25.70 |

Q64 joint readout reaches 74.32% macro accuracy and answers 3,764 of 5,112 items correctly, for 73.63% pooled accuracy. Its macro gain over the adapter-only Q64 parent is 7.86 percentage points. Improvements occur in all four benchmark columns, but are uneven: ANLI-R3 increases by 23.42 points, contributing 74.48% of the overall macro gain. This concentration matters when extrapolating to other decision workloads.

On paired items, joint Q64 corrects 551 errors made by its parent while losing 164 previously correct answers. The resulting 387 net correct answers correspond to a pooled gain of 7.57 points, which differs from the benchmark-weighted macro gain. Table 4 reports post-hoc macro intervals for the main paired comparisons. The parent gain is supported on this fixed collection; generalization to a new task mixture or a new training run remains unmeasured.

Table 4. Paired Q64 joint-readout differences. Positive values favor joint Q64. Intervals are post-hoc item-resampling intervals, conditional on the models and task strata. Wins and losses count items rather than macro-weighted observations.

| Comparison | Macro difference (pp) | 95% interval (pp) | Wins / losses |
|---|---|---|---|
| Q64 adapter only | +7.86 | [+6.86, +8.85] | 551 / 164 |
| Q32 adapter only | +13.52 | [+12.22, +14.81] | 992 / 298 |
| Native 27B | -1.37 | [-2.28, -0.46] | 258 / 335 |
| AnyJev Qwen3-32B L0 | -2.37 | [-3.52, -1.21] | 420 / 552 |

Joint Q64 remains 1.37 macro points below the native 27B system and 2.37 below AnyJev. ANLI-R3 is the exception: Q64 obtains 68.42%, versus 65.00% for native 27B and 65.08% for AnyJev. ARC, MMLU-Pro, and CMMLU remain lower. The evidence does not support overall superiority or accuracy equivalence. Similarly, the higher scores of deeper historical branches do not identify depth as the only explanation because data, initialization, and update scope changed across versions.

## 6.2 Request latency and amortization

Table 5. Complete-request median latency in milliseconds, nine repetitions per configuration and eight candidates per question. Cold excludes loading and CPU prompt preparation but includes document encoding; warm reuses document state. Accuracy was measured on a separate workload.

| Tokens / Q | Native cold | AnyJev serial | AnyJev batch | Q64 cold | Q64 warm |
|---|---|---|---|---|---|
| 256 / 1 | 177.98 | 186.10 | 185.24 | 245.85 | 86.18 |
| 256 / 10 | 1,007.75 | 1,860.43 | 1,519.90 | 398.89 | 237.97 |
| 256 / 50 | 4,685.92 | 9,294.69 | 7,598.37 | 1,379.96 | 1,220.27 |
| 4096 / 1 | 826.39 | 1,127.41 | 1,129.71 | 906.31 | 86.03 |
| 4096 / 10 | 1,650.55 | 11,284.82 | 11,351.26 | 1,100.46 | 280.14 |
| 4096 / 50 | 5,382.89 | 56,595.33 | 56,872.03 | 2,236.79 | 1,415.51 |

At 4,096 document tokens and 50 questions, Q64 takes 2.237 seconds cold and 1.416 seconds warm. The recorded native baseline takes 5.383 seconds, and batched AnyJev takes 56.872 seconds. Ratios of these cold medians are 2.41 and 25.43, respectively, relative to Q64. These are ratios between specific configurations, not per-token algorithmic constants or production throughput measurements. AnyJev's complete rotations and per-question document processing in the measured interface contribute work absent from Q64's single scoring path.

At 256 tokens and one question, cold Q64 takes 245.85 ms, compared with 177.98 ms for native 27B and 185.24 ms for batched AnyJev: Q64 is 38.13% and 32.72% slower, respectively. At 4,096 tokens and one question, Q64 is also slower than native 27B, although faster than the measured AnyJev setting. Document preparation is therefore an overhead when reuse opportunities are limited.

Joint Q64 and its adapter-only parent use the same inference structure. Across the six cold configurations, their median relative differences range from -1.03% to +0.42%. This observation supports an accuracy improvement without an added inference module, but it is not a formal latency-equivalence test. The nine repetitions are technical timing replicates, not independent devices or training runs.

Batching reduces short multi-question AnyJev latency by about 18% relative to its serial record, but makes little difference for long inputs under the tested memory limit. A parity check gives identical top choices on 32 real items and all 122 synthetic questions. The maximum real-item probability difference is 0.07703, so this is top-choice agreement on a small sample, not probability equivalence. The 76.68% accuracy in Table 3 belongs to the full serial evaluation; it is not a new full-set batched accuracy measurement.

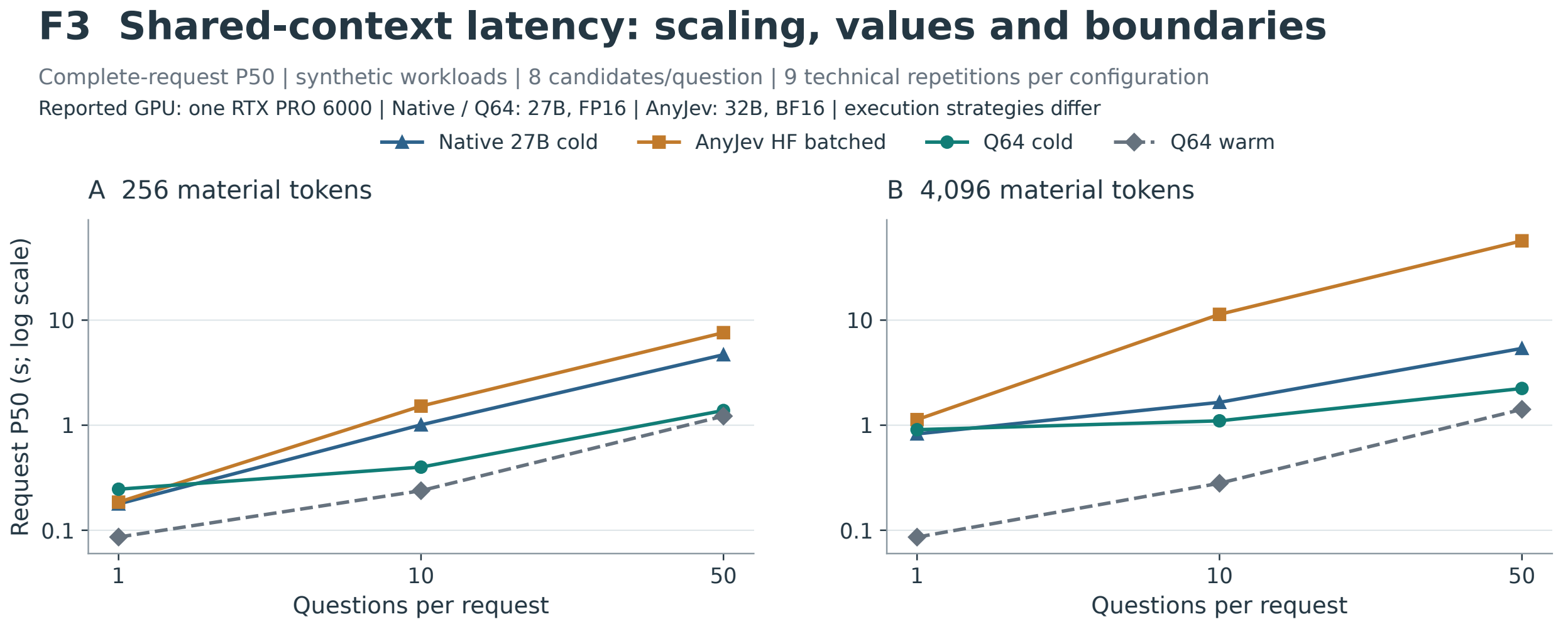

**C All six configurations: median complete-request time (s)**

| Tokens | Questions | Native cold | AnyJev batched | Q64 cold | Q64 warm |
|---|---|---|---|---|---|
| 256 | 1 | 0.178 | 0.185 | 0.246 | 0.086 |
| 256 | 10 | 1.008 | 1.520 | 0.399 | 0.238 |
| 256 | 50 | 4.686 | 7.598 | 1.380 | 1.220 |
| 4096 | 1 | 0.826 | 1.130 | 0.906 | 0.086 |
| 4096 | 10 | 1.651 | 11.351 | 1.100 | 0.280 |
| 4096 | 50 | 5.383 | 56.872 | 2.237 | 1.416 |

**D Effects and trade-off**

**4,096 tokens / 50 questions**
Native / Q64 cold: 2.41×
AnyJev / Q64 cold: 25.43×

**256 tokens / 1 question:**
Q64 cold 245.85 ms >
Native cold 177.98 ms

Cold includes material preparation; models and GPU kernels are already loaded/warm. Warm Q64 reuses material state.
Timing excludes model loading, CPU tokenization and prompt preparation; includes CPU probability return. Systems differ.

Discrete configurations: lines guide the eye. Source: latency_main.csv (ms ÷ 1,000); raw repetitions and serial baselines included.

Figure 3. Median complete-request latency for the separate synthetic workload: (A, B) 256-token and 4,096-token documents, (C) all six measured configurations in seconds, and (D) observed latency ratios and short-request cold overhead. Cold requests include document preparation; warm Q64 requests reuse it. Each point summarizes nine technical repetitions. Connecting lines guide the eye between the three measured question counts. Different backbones, precision, rotations, and batching policies limit causal attribution.

# 7 Discussion and limitations

The results suggest that rich frozen question representations and their trainable readout should be considered jointly. Updating the readout can recover accuracy lost by a mismatched interface while preserving the deployed computation graph. However, the final training stage broadens coverage as well as the parameter update set. A controlled experiment holding training rows, initialization, optimization budget, and seed constant would be needed to distinguish an adapter effect from a decoder, memory-projection, or candidate-reader effect.

The measured latency advantage is tied to amortization. Reusing document features and prefix states avoids repeating substantial computation, while grouping independent branches improves utilization. The native comparison already has a prefix cache, so the difference cannot be described simply as caching versus no caching. A native model with matched cross-question batching, identical precision, and the same serving engine may narrow the gap. The study does not evaluate vLLM or other AnyJev backends and does not transplant Q64 onto its 32B backbone.

Two generalization gaps are central. First, the public evaluation was repeatedly visible during research, and exact-overlap exclusions do not establish a clean semantic or pretraining separation. Second, actual multi-question documents have heterogeneous question lengths, candidate counts, and evidence requirements. The controlled latency workload does not measure those distributions. A newly sealed evaluation that measures correctness and latency on the same real-document sessions is the next decisive test.

The reader's attention mechanism is not a certificate of evidence use. The final stage uses only answer supervision, even though its pool contains factual-flip and irrelevant-change variants. Controlled removal or substitution of relevant evidence, candidate permutations, and calibrated abstention tests are needed to evaluate evidence sensitivity. Softmax scores should not be interpreted as reliable confidence without those checks.

Only one final-stage seed and one GPU configuration are represented. Item-bootstrap intervals are conditional statistical summaries and cannot repair development feedback or quantify cross-seed stability. Long-run service capacity, concurrency, peak memory under matched workloads, energy use, and quantized deployment remain unmeasured. The system also retains the full 27B backbone, so small incremental weights should not be mistaken for a lightweight deployment footprint.

## 8 Conclusion

Q64 combines reusable hybrid document states with full-depth question continuation and a jointly adapted candidate readout. In the recorded experiment, the final adaptation improves macro accuracy from 66.46% to 74.32% without adding an inference module. On a separate controlled long-document, many-question workload, its measured request latency is lower than the evaluated native and AnyJev configurations. Overall accuracy still trails both main comparisons, and cold single-question overhead remains visible. These results motivate further study of shared-context decision interfaces, with matched deployment baselines and fresh real-document evaluation required to establish broader benefits.

## 9 Artifact provenance and reproducibility

The manuscript is based on the supplied presentation, report archive, and selected-weight archive. The audit verifies 85 report-package file hashes; reaggregates predictions and timing repetitions; and checks both safetensors payload identities against the checkpoint manifest. All saved tensor values are finite, and the selected identity matches step 512 and the recorded development criterion. These checks establish consistency of the provided artifacts, not a new reproduction of model training or inference.

The accompanying analysis script and derived tables support reanalysis on a CPU with Python and NumPy. Model execution additionally requires the matching base weights, tokenizer, original runtime dependencies, historical training pools for retraining, and a suitable accelerator. The source snapshots retain imports from the original experiment environment and are not a complete standalone inference package. No public repository or DOI is claimed by this local draft.

Table 6. Recorded identities and verification scope. Full SHA-256 values are in the accompanying evidence manifest.

| Object | Identity prefix | What was checked |
|---|---|---|
| Presentation | 6d89dd48500d | File bytes and slide content |
| Report archive | b0c3dadb5524 | File bytes and 85 payload hashes |
| Selected weights archive | cd40f405daed | Tar payloads and tensor headers |
| Normalized evaluation set | c95c4d2ea18e | Recorded source identity only |
| Joint-stage protocol | b707877e3050 | Bytes match selected checkpoint identity |

The normalized evaluation dataset itself is not present in full text, so its recorded hash is a reconstruction target rather than a hash independently recomputed from this package. Source locations, per-model prediction records, raw timing arrays, and the post-hoc analysis seed are retained in the review materials. The selected incremental weights were inspected locally and are not duplicated into the manuscript source archive.

Manuscript preparation and post-hoc artifact analysis were assisted by OpenAI Codex on 29 October 2026. This preparation did not run new model training or inference. Affiliations, detailed contributions, funding, data-redistribution decisions, and final approval remain to be supplied by the researchers before public submission.